\documentclass{kaiart}
\usepackage{marvosym}
\usepackage{amsmath,amsfonts}
\usepackage{algorithmic}
\usepackage{array}
\usepackage{textcomp}
\usepackage{stfloats}
\usepackage{url}
\usepackage{verbatim}
\usepackage{graphicx, overpic, subfigure}
\usepackage{enumitem}
\usepackage{booktabs}
\usepackage{tabularx}
\usepackage{multirow}
\usepackage{helvet}

\usepackage{pifont}
\newcommand{\cmark}{\ding{51}}  % ✓
\newcommand{\xmark}{\ding{55}}

\title{Open-Vocabulary Camouflaged Object Segmentation with Cascaded Vision Language Models}
\author{
    Kai Zhao\textsuperscript{1,2},
    Wubang Yuan\textsuperscript{1},
    Zheng Wang\textsuperscript{1},
    Guanyi Li\textsuperscript{1}, \\
    Xiaoqiang Zhu\textsuperscript{1}$^{\text{\Letter}}$,
    Deng-Ping Fan\textsuperscript{3},
    Dan Zeng\textsuperscript{1}
}
\affiliation{
    \vspace{0.5em}
    \textsuperscript{1} Shanghai University \ \
    \textsuperscript{2} UCLA \ \ 
    \textsuperscript{3} Nankai University \\
    \vspace{0.5em}
    \texttt{kz@kaizhao.net, fdp@nankai.edu.cn \\ \{yuanwubang, zhengwang, kwunyatlee, xqzhu, dzeng\}@shu.edu.cn} \\
}

\begin{document}
\maketitle

\begin{abstract}
Open-Vocabulary Camouflaged Object Segmentation (OVCOS) seeks to segment and classify camouflaged objects from arbitrary categories, presenting unique challenges due to visual ambiguity and unseen categories.
Recent approaches typically adopt a two-stage paradigm: first segmenting objects, then classifying the segmented regions using Vision Language Models (VLMs).
However, these methods (1) suffer from a domain gap caused by the mismatch between VLMs' full-image training and cropped-region inference, 
and (2) depend on generic segmentation models optimized for well-delineated objects, making them less effective for camouflaged objects.
Without explicit guidance, generic segmentation models often overlook subtle boundaries, 
leading to imprecise segmentation.
In this paper,
we introduce a novel VLM-guided cascaded framework 
to address these issues in OVCOS.
For segmentation, 
we leverage the Segment Anything Model (SAM), 
guided by the VLM.
Our framework uses VLM-derived features as explicit prompts to SAM, effectively directing attention to camouflaged regions and significantly improving localization accuracy.
For classification, 
we avoid the domain gap introduced by \emph{hard} cropping.
Instead, we treat the segmentation output as a \emph{soft} spatial prior via the alpha channel, 
which retains the full image context while providing precise spatial guidance, leading to more accurate and context-aware classification of camouflaged objects.
The same VLM is shared across both segmentation and classification to ensure efficiency and semantic consistency.
Extensive experiments on both OVCOS and conventional camouflaged object segmentation benchmarks demonstrate the clear superiority of our method, highlighting the effectiveness of leveraging rich VLM semantics for both segmentation and classification of camouflaged objects.
The code and models are open-sourced at \url{https://github.com/intcomp/camouflaged-vlm}.
\end{abstract}

\section{Introduction}

Open-Vocabulary Camouflaged Object Segmentation (OVCOS) is a challenging task that requires segmenting and classifying camouflaged objects of novel categories not seen during training~\cite{r1}. Compared to traditional semantic segmentation~\cite{maskformer, NEURIPS2021_64f1f27b, Chen_2018_ECCV}, OVCOS faces additional challenges because it requires recognizing novel categories in visually ambiguous scenes, where camouflage leads to low contrast, indistinct boundaries, and high similarity between objects and their backgrounds. These challenges are particularly relevant in real-world applications such as medical image analysis~\cite{sinet} and agricultural monitoring~\cite{liu2019pestnet}, where annotations are scarce and target categories are often seen. 

Several existing open-vocabulary segmentation approaches~\cite{r4, lseg, bucher2019zero, xian2019semantic} utilize vision-language models (VLMs), e,g, CLIP~\cite{r13},
to directly classify each pixel across the entire input image, thereby improving semantic generalization. 
These approaches operate under a one-stage framework.
However, VLMs are pre-trained for image-level understanding,
creating a granularity mismatch that hinders effective visual-semantic alignment and limits semantic transfer, often leading to suboptimal performance~\cite{r7}.

\begin{figure}[tb]
    \centering
    \subfigure[Generic Segmentation Model for COS]{
    \includegraphics[width=0.5\linewidth]{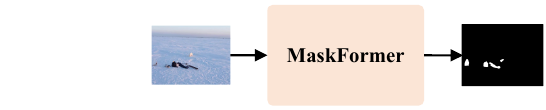}
    \label{fig:seg_sub_fig1}
    }
    \vspace{-0.5em} % 设置子图之间的垂直间距
    \subfigure[Visual-Language Prompted for COS]{
    \includegraphics[width=0.5\linewidth]{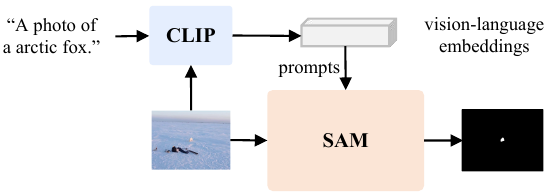}
    \label{fig:seg_sub_fig2}}
    \vspace{-0.5em}
    
    \caption{
    Different Camouflaged Object Segmentation (COS) paradigms in two-stage OVCOS.
    (a) Generic segmentation models, such as MaskFormer~\cite{maskformer}, typically operate directly on the input image without target-specific guidance, and are primarily designed to segment salient foreground objects. (b) Our segmentation model leverages vision-language embeddings from CLIP as prompts to guide the SAM model, directing attention to the camouflaged area.}
    \label{fig:segmentation_compare}
\end{figure}

To mitigate this gap, recent works~\cite{r1, r7,zegformer, maskclip, r5} first perform class-agnostic segmentation and then classify the segmented regions using VLM. 
This pipeline forms a two-stage framework. 
This decoupling of segmentation and classification partially alleviates the granularity mismatch~\cite{r7}. 
However,
in the segmentation stage, 
as shown in Figure~\ref{fig:seg_sub_fig1}, many existing approaches typically rely on the generic segmentation architectures~\cite{r7, zegformer, maskclip, r5, maskformer}
to spot the target region.
These generic segmentation models are primarily tailored for well-delineated objects and might fail to generalize effectively to camouflaged scenarios,
where targets are subtle, indistinct, and visually embedded in complex backgrounds. 
The lack of alignment between the pretraining objectives and the demands of camouflaged segmentation leads to imprecise localization. 
In addition, most existing methods do not incorporate explicit edge-aware mechanisms, 
which are crucial for accurately delineating objects with weak or ambiguous boundaries.

%
% This absence of semantic conditioning limits segmentation performance, especially in camouflaged scenarios that require fine-grained discrimination. In addition, most methods lack explicit edge-aware modeling, which is essential to accurately delineate objects with subtle boundary cues.

Recent advances in foundation models such as the Segment Anything Model (SAM)~\cite{r20} have shown remarkable generalization across various segmentation tasks, largely due to their ability to perform prompt-guided segmentation. 
By using prompts to specify target regions, SAM can adapt its attention to user-defined areas, making it particularly effective for specialized tasks such as camouflaged object segmentation.
We propose an adapted SAM architecture tailored for camouflaged object segmentation.
As illustrated in ~\cref{fig:seg_sub_fig2}, we integrate CLIP-derived visual and textual embeddings as prompts into the SAM mask decoder, providing task-specific semantic guidance that enhances the ability of the model to focus on the camouflaged targets. 
\begin{figure}[!htb]
    \centering
    \subfigure[\emph{Hard} guidance via Mask Cropping]{
    \includegraphics[width=0.5\hsize]{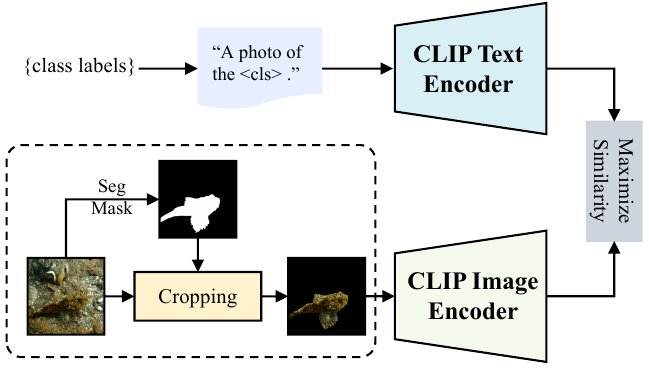}
    \label{fig: cls_sub_fig1}}
    
    \vspace{-0.5em} % 设置子图之间的垂直间距

    \subfigure[Our \emph{Soft} spatial guidance]{
    \includegraphics[width=0.5\hsize]{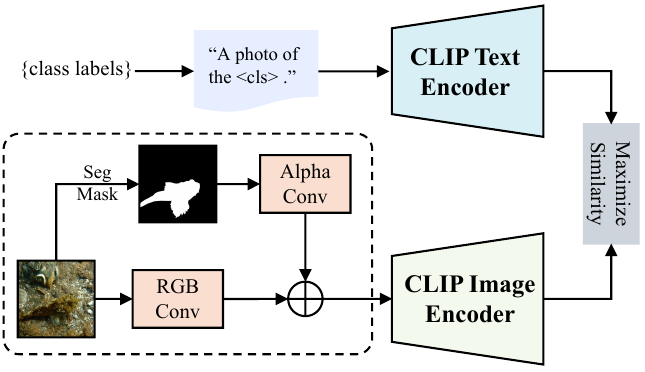}
    \label{fig: cls_sub_fig2}}
    \vspace{-0.5em}
    \caption{Comparison of mask-guided classification strategies. (a) Mask cropping strategy: applies the segmentation mask to crop the input image before feeding it into the CLIP image encoder. (b) Ours: fuses the segmentation mask with the original image for region-aware classification while retaining full-image context.
}
    \label{fig: classification_compare}
\end{figure}
Additionally, we enhance the mask decoder with conditional multi-way attention and an edge-aware refinement module to improve boundary precision, effectively handling the indistinct contours characteristic of camouflage. 

In the classification stage, most existing methods crop the segmented regions for classification~\cite{r1, r7, zegformer} (Figure~\ref{fig: cls_sub_fig1}), introducing a domain gap since CLIP is pre-trained on full images. 
%
%Others apply learnable prompts to only the visual or textual modality~\cite{maskclip, r5} (Figure~\ref{fig:b}), limiting the use of CLIP multi-modal capabilities and weakening alignment and generalization. 
%
To mitigate the domain gap, 
we adopt a region-aware classification strategy that replaces hard cropping with a soft spatial prior derived from the segmentation mask, applied via the image’s alpha channel.
Our approach preserves the full image context while providing explicit spatial guidance.
The predicted segmentation mask serves as a soft spatial prior and is fused with the input image via a lightweight integration module before being processed by the CLIP~\cite{r13} image encoder. 
~\cref{fig: classification_compare} compares
the \emph{hard} and \emph{soft}
spatial guidance.
Additionally, we fine-tune CLIP using a multi-modal prompting strategy similar to~\cite{r10}, jointly optimizing both visual and textual prompts. 
This enhances semantic alignment and task-specific adaptability, enabling region-aware classification without disrupting global semantics.

Building on these ingredients, we introduce the \textbf{C}ascaded \textbf{O}pen-vocabulary \textbf{C}amouflaged \textbf{U}nder\textbf{S}tanding network (COCUS), a novel two-stage framework for the OVCOS task that explicitly decouples the process into \textit{segment and classify}.
In the first stage (segmentation), 
we use CLIP~\cite{r13} to extract visual and textual features.
These features serve as prompts to the SAM~\cite{r20} for segmentation. 
This prompt-based guidance allows SAM to focus more precisely on camouflaged target regions, 
enhancing localization in visually ambiguous scenes. 
In the second stage (classification), the segmentation output serves as spatial guidance to refine the integration with the original image, allowing CLIP to perform open-vocabulary classification with improved focus on target regions.
%
%By disentangling segmentation and classification, COCUS delivers fine-grained semantic understanding of camouflaged objects and overcomes key limitations of prior segmentation-guided pipelines.
By disentangling segmentation and classification, our method enables more accurate semantic interpretation of camouflaged objects through prompt-based guidance segmentation and region-aware classification.

Extensive experiments on the OVCamo~\cite{r1} benchmark demonstrate the effectiveness of the proposed framework for the OVCOS task. 
Compared to the strong baseline OVCoser~\cite{r1}, we achieve consistent improvements across all major evaluation metrics, establishing a new state-of-the-art on this challenging benchmark. 
Moreover, the adapted SAM~\cite{r20} demonstrates strong performance on the conventional COS task, confirming that CLIP-derived embeddings prompting and edge-aware refinement remains effective in standard closed-set scenarios.

\vspace{0.5em}
The main contributions of this work are as follows:
\begin{itemize}
\item We propose a novel two-stage framework for OVCOS that explicitly decouples segmentation and classification. 
Our approach employs a prompt-guided segmentation model to generate a mask,
which serves as a \emph{soft} spatial guidance for the classification stage
while preserving full-image context.

\item We propose an adapted SAM as the segmentation model, 
enhanced for camouflaged object localization by injecting CLIP-derived textual and visual embeddings as prompts.
This design provides rich semantic guidance that steers attention toward visually ambiguous regions.
Furthermore, we improve SAM's mask decoder with conditional multi-way attention and edge-aware refinement, 
improving both spatial accuracy and boundary delineation.

\item Extensive experiments on the OVCamo benchmark demonstrate that our method achieves state-of-the-art performance. Moreover, the adapted SAM exhibits strong generalization on the conventional COS task, validating the effectiveness of our framework across both open- and closed-set camouflaged segmentation scenarios.

\end{itemize}

The remainder of this paper is organized as follows. Section~\ref{sec:related_work} reviews recent advances in open-vocabulary segmentation and camouflaged object understanding. Section~\ref{sec:method} presents the proposed framework, detailing its cascaded design, CLIP fine-tuning pipeline and adapted SAM segmentation model. Section~\ref{sec:experiments} presents implementation details, including training settings and architectural configurations, followed by comprehensive experimental results and ablation studies.
\section{Related Work}
\label{sec:related_work}
\subsection{Vision-Language Models}
Vision-language models (VLMs) are neural architectures that learn joint visual-textual representations by embedding both image and text inputs into a shared semantic space.
A seminal model in this domain, CLIP~\cite{r13}, jointly learns image and text representations via contrastive learning on large-scale web data, demonstrating strong generalization across open-vocabulary tasks such as object detection~\cite{zang2022open, gu2021open, zareian2021open, li2023opensd} and segmentation~\cite{r4,r7, zegformer,maskclip,r5,r6, r12, segclip, 122learning}. However, vanilla CLIP often performs poorly in downstream tasks without task-specific adaptation. To address this limitation, researchers have proposed a variety of fine-tuning approaches. Alpha-CLIP~\cite{r21} introduces spatially adaptive attention to enhance focus on semantically relevant image regions. CoOp~\cite{zhou2022learning} and Co-CoOp~\cite{cocoop} optimize textual prompts for better few-shot performance and generalization, respectively. Visual prompt tuning~\cite{bahng2022visual} further enhances adaptability by injecting fine-grained prompts into the vision branch. To overcome the limitations of single-modality tuning, recent works~\cite{r10,khattak2023self, fgvp} adopt multi-modal strategies. FGVP~\cite{fgvp} learns patch-level visual prompts to improve alignment across diverse tasks. MaPLe~\cite{r10} jointly tunes prompts in both visual and textual encoders, preserving CLIP’s generality while enabling task-specific adaptation. In this work, we adopt a multi-modal prompt tuning framework similar to MaPLe to fine-tune CLIP, enhancing semantic alignment for OVCOS. 

\subsection{Camouflaged Object Segmentation} 
Camouflaged Object Segmentation (COS) has emerged as a significant research focus in computer vision, with the aim of segmenting objects that visually blend into their surroundings. Unlike traditional tasks such as salient object detection~\cite{borji2019salient, pang2020multi, ji2022dmra, liu2022poolnet+, li2023delving} and semantic segmentation~\cite{ji2023multispectral, zhao2021multi}, COS is inherently more challenging than traditional segmentation tasks due to low object-background contrast, ambiguous boundaries, and high background similarity. It holds practical value in domains such as medical image analysis~\cite{sinet} and agricultural monitoring~\cite{liu2019pestnet}. COS is typically formulated as a class-agnostic task, focusing on segmenting camouflaged regions within complex visual scenes. The available work~\cite{sinet, sinet-0,nc4k, pfnet,  zoomnet, segmar, dgnet} to date has demonstrated strong performance on established benchmark datasets~\cite{sinet, nc4k,camo}. Recent advances have introduced several SAM-based methods~\cite{r20, samadapter, sammed} adapted for COS, which use prompt tuning and architectural modifications to improve segmentation performance in complex scenes. 

\subsection{Open-Vocabulary Camouflaged Object Segmentation} 

Open-Vocabulary Camouflaged Object Segmentation (OVCOS) is a specialized subtask of open-vocabulary segmentation, in which the goal is to segment and recognize camouflaged objects belonging to arbitrary textual categories. Open-vocabulary segmentation aims to align visual and textual representations in a shared embedding space, enabling pixel-level segmentation for unseen or novel categories. Early methods~\cite{zhao2017open} used semantic hierarchies and concept graphs to bridge word concepts and semantic relations. With the rise of VLMs like CLIP~\cite{r13}, recent open-vocabulary segmentation methods have shifted toward leveraging pretrained VLMs to directly connect visual regions with text queries. These approaches follow one-stage and two-stage paradigms. One-stage methods such as MaskCLIP~\cite{maskclip} adapt CLIP for segmentation without additional training. SAN~\cite{r6} enhances feature representations via adapters. CAT-Seg~\cite{r4} introduces cost aggregation between image and text embeddings. And FC-CLIP~\cite{r12} employs hierarchical feature fusion. However, these methods often suffer from suboptimal alignment due to CLIP's image-level representations. The two-stage methods address this by decoupling segmentation and classification. For example, SimSeg~\cite{r7} uses a cascaded design with MaskFormer~\cite{maskformer} for class-agnostic mask generation and CLIP for classification. OVSeg~\cite{r5} fine-tunes CLIP on diverse and noisy data to improve generalization. In~\cite{r11}, text-to-image diffusion model is employed for mask generation. While these two-stage framework methods work well on generic objects, they fall short in camouflaged scenarios. OVCOS is especially difficult because low contrast visuals, ambiguous edges, and visually similar backgrounds all contribute to degraded segmentation and classification results. OVCoser~\cite{r1} is the first to address this task by combining a dedicated camouflaged segmentation model with a CLIP-based classifier in a two-stage pipeline. However, it relies on cropped inputs for classification and does not fully exploit VLM semantics in segmentation. 
\section{Methodology}\label{sec:method}
\begin{figure}[!htb]
  \centering
  \includegraphics[width=0.5\linewidth]{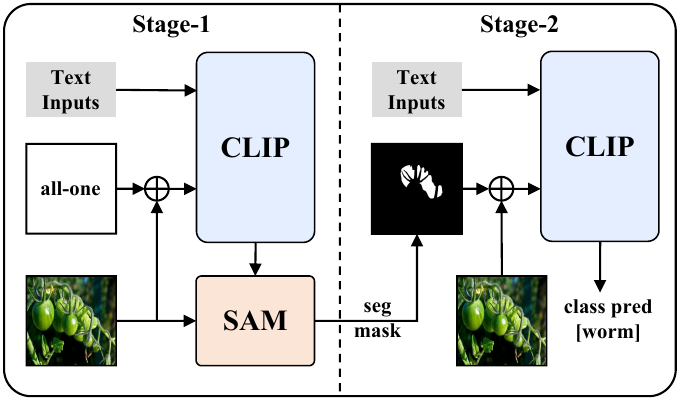}
    \caption{Overview of the cascaded \textit{segment and classify} framework. In Stage-1, the adapted SAM model generates a class-agnostic camouflaged segmentation mask using textual and visual embeddings as prompts. In Stage-2, we use the generated segmentation mask to enable region-aware open-vocabulary classification.}
  \label{fig:Overview}
\end{figure}
\subsection{Problem Definition}
Open-Vocabulary Camouflaged Object Segmentation (OVCOS) aims to \textit{segment and classify} camouflaged objects belonging to novel categories unseen during training. Formally, let \(\mathcal{C}_{\text{seen}}\) denote the set of categories available during training, and \(\mathcal{C}_{\text{unseen}}\) represent the disjoint set of target categories at inference time, such that \(\mathcal{C}_{\text{seen}} \cap \mathcal{C}_{\text{unseen}} = \emptyset\). Given an input image \(I\) and novel class labels \(\mathcal{C}_{\text{unseen}}\), the model is required to produce a segmentation mask \(M\) highlighting the camouflaged object and predicts its corresponding class label \(\hat{y} \in \mathcal{C}_{\text{unseen}}\).

To address this task, we adopt a \emph{segment-and-classify} strategy. In the first stage, a class-agnostic segmentation model localizes camouflaged regions guided by visual and textual semantics. In the second stage, a vision-language model performs open-vocabulary classification by comparing the visual representation of the segmented regions with textual embeddings of the novel class labels, supporting recognition in an open-set setting.

\subsection{Overview}
~\cref{fig:Overview}
demonstrates
the proposed two-stage framework for the OVCOS.
During inference, the first stage generates a class-agnostic camouflaged segmentation mask, while the second stage performs open-vocabulary classification based on the segmented regions.
We use the same CLIP model for both stages.
Our CLIP model accepts
a triplet \{$I \in \mathbb{R}^{H \times W \times 3}, M$, text\} as input,
where $I$ and $M$ are image and mask,
and text is a description of the input,
with the format of `a photo of \(<\emph{something}>\)'.
The CLIP model outputs visual and textual embeddings, 
$E_v$ and $E_t$, 
which serve as prompts to guide segmentation in the first stage and are used for similarity-based open-vocabulary classification in the second stage.
Notably, to ensure a consistent input format across stages, we use an all-one mask in the first stage, while in the second stage, the predicted segmentation mask is used as input.

In the first stage, 
as shown in ~\cref{fig:Overview} (left),
we perform segmentation guided by textual and visual embeddings. The inputs consist of an RGB image \( I \in \mathbb{R}^{H \times W \times 3} \)
% , an alpha mask \( A \in \mathbb{R}^{H \times W \times 1} \) filled with all entries set to one, 
and a set of class labels \(\mathcal{C} = \{c_1, c_2, \dots, c_N\}\), where \( N \) denotes the number of candidate classes.
They are processed by the CLIP~\cite{r13} model to produce textual embedding \(E_t\) and visual embedding \(E_v\) optimized for camouflaged object understanding. These embeddings serve as prompts and, together with the image \( I \), are input into the adapted SAM model to guide the prediction of a class-agnostic camouflaged segmentation mask \( M \in [0,1]^{H \times W \times 1} \), effectively localizing the camouflaged object.

As shown in~\cref{fig:Overview} (right),
in the second stage, we perform open-vocabulary classification guided by the segmented result. The inputs include the same RGB image \( I \) and class labels \( \mathcal{C}_{\text{unseen}} \) from the first stage,
and the predicted segmentation mask \( M \)
which is used as an additional input
to the CLIP model as spatial guidance.
These inputs are processed by the CLIP model
as stage one,
which now focuses more precisely on the localized object area.
The model then outputs a predicted class label \( \hat{y} \in \mathcal{C}_{\text{unseen}} \), identifying the category of the camouflaged object.
Let $E_t^N\in\mathbb{R}^{N\times d}$,
and $E_v \in\mathbb{R}^{1\times d}$
be the textual and visual embeddings,
where $d=768$ is the feature dimension,
we first calculate the similarity scores $S\in\mathbb{R}^N$:
\begin{equation}
    S = E_t^N\cdot (E_v)^T.
    \label{eq:sim}
\end{equation}

During training, we first fine-tune our CLIP~\cite{r13} model by optimizing learnable prompts in both the language and vision branches to enhance its sensitivity to camouflaged objects, while keeping all encoder parameters fixed.
~\cref{fig:train_first_stage}
illustrates the fine-tuning pipeline of our CLIP.
After fine-tuning, we freeze the CLIP model as a feature extractor and train the SAM~\cite{r20} model using visual-textual features from CLIP as prompts. The details of the CLIP fine-tuning process are provided in Section~\ref{sec:tuned-alpha-clip}, and the architecture of the adapted SAM is described in Section~\ref{sec:adapted-sam}.
%
% by incorporating cross-modal embeddings guidance and an edge-enhanced mask decoder to produce precise object masks and boundaries. 

\subsection{CLIP Fine-Tuning Pipeline}
\label{sec:tuned-alpha-clip}

\begin{figure}[!htb]
  \centering
  \includegraphics[scale=0.67]{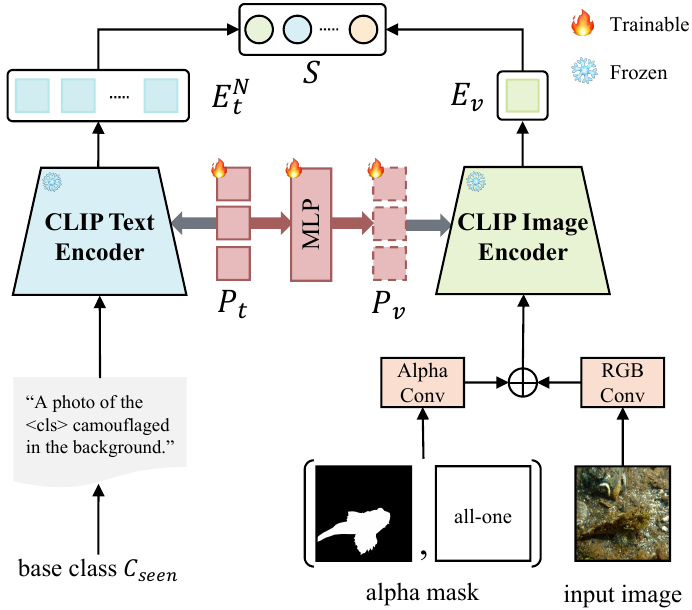}
    \caption{The CLIP fine-tuning pipeline. The language branch encodes base class labels \(C_{seen}\) with a camouflage-specific prompt template and learnable textual prompts \( P_t \) to obtain textual embeddings \( E_t^N \). The vision branch fuses features from the input image and alpha mask, combined with visual prompts \( P_v \) injected via an MLP, and passes them to the frozen CLIP image encoder to obtain visual embedding \( E_v \). Similarity scores \( S \) are computed by aligning \( E_t^N \) and \( E_v \) in a shared space.} 
  \label{fig:train_first_stage}
\end{figure}

We fine-tune the CLIP model using a multi-modal prompting strategy to enhance its ability to capture subtle semantic cues for camouflaged object segmentation, as shown in Figure~\ref{fig:train_first_stage}. Our CLIP variant is a modified version of Alpha-CLIP~\cite{r21}.
Previous prompting strategies in CLIP~\cite{r13} typically operate on visual or textual modality. Language-only prompt tuning methods~\cite{r7,zhou2022learning, cocoop} optimize learnable prompts solely in the language branch, while visual-only approaches~\cite{maskclip, r5, fgvp} inject prompts exclusively into the vision branch. In this work, we adopt a multi-modal prompting strategy, following~\cite{r10}, which jointly optimizes both textual and visual prompts to enhance multi-modal alignment and better adapt to task-specific objectives. 

Specifically, we append learnable textual prompts \(P_t\) to the language branch and generate the corresponding visual prompts \(P_v\), which are produced by conditioning on the textual prompts through an MLP injector. 
The language and textual prompts $P_t$ and $P_v$
are shown in middle of~\cref{fig:train_first_stage}.
During fine-tuning, only the textual prompts and injector parameters are updated, while the rest of CLIP model remains frozen. This lightweight strategy promotes efficient adaptation and enables improved semantic alignment across modalities. Below, we outline the fine-tuning pipeline of CLIP model.

% \subsubsection{Training Pipeline}
The fine-tuning pipeline begins with the language branch, where the base class labels $\mathcal{C_\text{seen}}$ are formatted using the prompt template ``A photo of the $<class>$ camouflaged in the background.'' and enriched with learnable textual prompts \( P_t \). These are processed by the frozen CLIP text encoder to produce textual embeddings \( E_t^{N} \in\mathbb{R}^{N \times 768} \), where \( N \) is the number of the base classes.

Currently, in the vision branch, the input RGB image \( I \in \mathbb{R}^{H \times W \times 3} \) is combined with an auxiliary alpha mask \( A \in \mathbb{R}^{H \times W \times 1} \). The alpha mask \( A \) is randomly selected as either the all-one mask \( A_J \) or the ground-truth segmentation mask \( A_{gt} \), each with equal probability.
This enables the CLIP model to optionally accept a mask as input, defaulting to an all-one matrix when no mask is provided—for example, during the first stage of segmentation.
% \begin{equation}
% A = 
% \begin{cases}
% A_{J}, & \text{with probability } 0.5 \\
% A_{gt}, & \text{with probability } 0.5
% \end{cases}
% \end{equation}

The image \( I \) and the alpha mask \( A \) are separately processed through dedicated convolutional layers,
e.g. AlphaConv and RGBConv in ~\cref{fig:train_first_stage},
to extract modality-specific features, which are then fused to form the visual representation. This fused representation, along with the injected visual prompts \( P_v \) generated by a lightweight MLP-based injector, is fed into the frozen CLIP image encoder to obtain the visual embedding \( E_v \in \mathbb{R}^{1 \times 768} \).

% Finally, the resulting textual embeddings and visual embedding are projected into a shared embedding space and compared using a similarity-based matching function to compute classification scores \(S \in \mathbb{R}^{N \times 1} \), thus effectively aligning vision and language representations.

Finally,
the textual and visual embeddings
are used to compute the similarity score as defined 
in~\cref{eq:sim},
which is used to calculate a cross-entropy loss
against the ground-truth class labels.

% Finally, the resulting textual embeddings \( E_t^{N}\) and visual embedding \( E_v \) are projected into a shared embedding space and compared using a similarity-based matching function to compute classification scores \( S  \). The model is trained using a cross entropy loss between the predicted scores and the ground-truth class label, encouraging accurate alignment of visual inputs with their corresponding textual categories.

\subsection{Adapted SAM}
\label{sec:adapted-sam}

\begin{figure*}[!htb]
  \centering
  \begin{overpic}[scale=0.53]{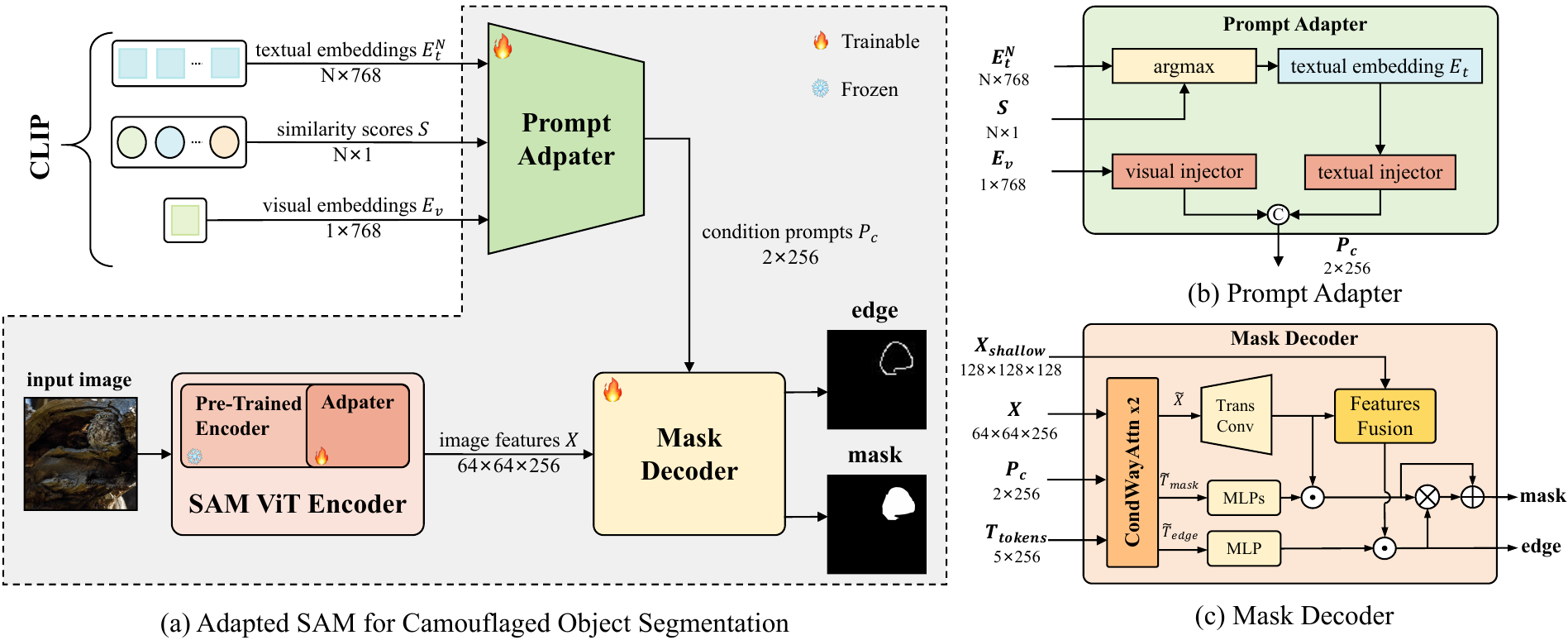}
  % \put(95, 5){edge}
  \end{overpic}
    \caption{
    \textbf{Overview of the adapted SAM framework.} 
\textbf{(a)} \textit{Adapted SAM for COS:} Our fine-tuned CLIP provides textual embeddings $E_t^N$, visual embedding $E_v$, and similarity scores $S$, which are projected into condition prompts $P_c$ via a Prompt Adapter. Image features $X$ extracted by SAM ViT encoder are refined by adapters. The Mask Decoder integrates $X$ and $P_c$ to predict the segmentation mask $M$ and edge map $E$, enabling precise localization. 
\textbf{(b)} \textit{Prompt Adapter:} Selects the most relevant textual embedding based on $S$, and projects both $E_t$ and $E_v$ into a unified condition space via lightweight MLPs to guide the decoder. 
\textbf{(c)} \textit{Adapted Mask Decoder:} Combines image features $X$, condition prompts $P_c$, and output tokens $T_{\text{tokens}}$ to produce accurate masks and edge maps, improving segmentation in camouflaged scenes.
    }
  \label{fig:train_second_stage}
\end{figure*}

We build upon the SAM~\cite{r20} to address the unique challenges of COS. While SAM excels at general-purpose segmentation, it struggles with the subtle visual cues and semantic ambiguities inherent to camouflaged objects. To overcome these limitations, as shown in Figure~\ref{fig:train_second_stage}\textcolor{blue}{(a)}, we adapt SAM by incorporating textual and visual embeddings guidance and edge-aware enhancements for improved segmentation.

Specifically, we integrate our fine-tuned CLIP model with SAM to provide semantic context. The CLIP model produces textual embeddings \( E_t^{N} \in \mathbb{R}^{N \times 768} \), a visual embedding \( E_v \in \mathbb{R}^{1 \times 768} \), and similarity scores \( S \in \mathbb{R}^{N \times 1} \). These embeddings are further processed by a prompt adapter, which projects them into condition prompts \( P_c \in \mathbb{R}^{2 \times 256} \), providing high-level semantic guidance into the segmentation pipeline.

In parallel, the SAM ViT encoder extracts image features  \( X \in \mathbb{R}^{64 \times 64 \times 256} \) from the input image. To adapt SAM to camouflage-specific cues, we introduce lightweight adapter modules that refine the image features \( X \) while keeping the backbone frozen.

Finally, the refined image features \( X \) and condition prompts \( P_c \) are fused within a mask decoder, which outputs a segmentation mask \( M \in \mathbb{R}^{H \times W \times 1} \) and an edge map \( E \in \mathbb{R}^{H \times W \times 1} \). The integration of refined image features and condition prompts within the decoder ensures accurate object localization and precise boundary delineation.

\subsubsection{Prompt Adapter}

The Prompt Adapter refines textual and visual embeddings from our fine-tuned CLIP to generate condition prompts for segmentation guidance, as shown in Figure~\ref{fig:train_second_stage}\textcolor{blue}{(b)}. Given textual embeddings \( E_t^N = \{ e_t^{1}, e_t^{2}, \dots, e_t^{N} \} \), visual embedding \( E_v \), and similarity scores \( S = \{ s_1, s_2, \dots, s_N \} \), the adapter selects the textual embedding corresponding to the highest similarity score:
\begin{equation}
i^* = \arg\max_{i} s_i, \quad E_t = e_t^{i^*}. 
\end{equation}

The selected textual embedding \( E_t \) and visual embedding \( E_v \) are projected into a shared 256-dimensional condition space using lightweight MLP-based injectors. The resulting condition prompts \( P_c \in \mathbb{R}^{2 \times 256} \) provide high-level semantic and visual guidance to the segmentation mask decoder, enhancing object localization and boundary accuracy. Formally, this is defined as:
\begin{equation}
P_t = \text{MLP}_{\text{text}}(E_t), \quad P_v = \text{MLP}_{\text{vis}}(E_v), 
\end{equation}
\begin{equation}
P_c = [P_t, P_v] \in \mathbb{R}^{2 \times 256}, 
\end{equation}
where \(\text{MLP}_{\text{text}}(\cdot)\) and \(\text{MLP}_{\text{vis}}(\cdot)\) denote the projection functions for textual and visual features, respectively.

\subsubsection{Mask Decoder}
We adapt the original SAM~\cite{r20} mask decoder to address the specific challenges of camouflaged object segmentation by introducing semantic conditioning and edge-aware enhancements. The modified decoder integrates multi-level image features \(X\), condition prompts \(P_c\), and output tokens \( T_{\text{tokens}} \), including mask tokens \( T_{\text{mask}} \) and an edge token \( T_{\text{edge}}\), to localize objects precisely and refine boundaries accurately, as shown in Figure~\ref{fig:train_second_stage}\textcolor{blue}{(c)}.

We first apply two Conditional Multi-Way Attention (CondWayAttn~$\times$2) modules to model the interactions among image features, condition prompts, and tokens. Each block enables dense bidirectional information flow between the image features, condition prompts, and output tokens. Specifically, it includes image-to-token and image-to-condition attention to incorporate visual context, token-to-condition and token-to-image attention to align output tokens with semantic and spatial cues, as well as token self-attention and an MLP layer to capture intra-token dependencies and perform feature transformation. The enhanced outputs are computed as:

\begin{equation}
\tilde{X}, \tilde{T}_\text{{mask}}, \tilde{T}_\text{{edge}} = CondWayAttn(X, P_c, T_\text{token}). 
\end{equation}

The attention-enhanced features \( \tilde{X} \) are first upsampled using a transposed convolution to restore spatial resolution. To recover fine-grained details, these features are fused with shallow image features \( X_{\text{shallow}} \) through a fusion block defined as:

\begin{equation}
\begin{aligned}
X_{\text{fusion}} =\ & \text{TConv}(\tilde{X})\ + \\
& \text{Conv} \big( \text{ReLU} \big( \text{Norm} \big( \text{Conv}(X_{\text{shallow}}) \big) \big) \big). 
\end{aligned}
\end{equation}

The attention-enhanced mask and edge tokens are then projected via task-specific MLPs. The coarse segmentation mask is computed by element-wise multiplication of the mask token with the upsampled features:
\begin{equation}
M_{\text{coarse}} = \text{MLPs}(\tilde{T}_\text{{mask}}) \odot \text{TConv}(\tilde{X}). 
\end{equation}

Similarly, the edge map is predicted by interacting the edge token with the fused feature map:
\begin{equation}
E = \text{MLP}(\tilde{T}_\text{{edge}}) \odot X_{\text{fusion}}. 
\end{equation}

The final refined mask incorporates edge guidance by multiplying the coarse mask with the edge map, followed by a residual addition:
\begin{equation}
M_{\text{fine}} = M_{\text{coarse}} + \left( M_{\text{coarse}} \otimes E \right). 
\end{equation}

This edge-guided refinement enhances boundary accuracy while preserving regional consistency, effectively handling low-contrast and subtle camouflaged structures.
The effectiveness of this module is evidenced by the ablation study in~\cref{sec:abl-mask-dec}.
\section{Experiments}
\label{sec:experiments}
\subsection{Implement Details}
\label{sec:implement_details}
\subsubsection{Datasets}
We evaluate our method on two tasks: Open-Vocabulary Camouflaged Object Segmentation (OVCOS) and Camouflaged Object Segmentation (COS).

For the OVCOS task, all experiments are conducted on the OVCamo~\cite{r1} dataset, a benchmark specifically curated for this setting. It comprises 11,483 images sourced from various publicly available datasets, covering 75 camouflaged object categories embedded in complex natural scenes. To enable open-vocabulary evaluation, the dataset is divided into two disjoint subsets by category: the training set \( \mathcal{D}_{\text{train}} \) includes 7,713 images from 14 seen categories, while the test set \( \mathcal{D}_{\text{test}} \) contains 3,770 images from 61 unseen categories, following an approximate 7:3 split.

For the COS task, we evaluate on three widely used benchmarks: CAMO~\cite{camo}, COD10K~\cite{sinet}, and NC4K~\cite{nc4k}. A total of 4,040 images from CAMO and COD10K are used for training. We conduct evaluation on the remaining images from these datasets, as well as the entire NC4K set. The detailed statistics of all datasets, covering training/testing splits, are presented in Table~\ref{tab:all-datasets}.

\begin{table}[!htb] 
    \centering
    \caption{Summary of datasets used for OVCOS and COS tasks.}
    \label{tab:all-datasets}
    \resizebox{0.6\linewidth}{!} % graphicx包的命令；\linewidth：表示当前文本区域的宽度；!：表示自动按比例缩放高度
    {
    \begin{tabular}{l|c|cccc}
        \toprule
        \textbf{Dataset} & \textbf{Task} & \textbf{Total} & \textbf{Train} & \textbf{Test} & \textbf{Categories} \\
        \midrule
        OVCamo~\cite{r1}          & OVCOS & 11,483 & 7,713 & 3,770 & 75 (14/61) \\
        CAMO~\cite{camo}            & COS   & 1,250  & 1,000 & 250   & -- \\
        % CHAMELEON~\cite{chamellon}       & COS   & 76     & --    & 76    & -- \\
        COD10K~\cite{sinet}          & COS   & 5,066  & 3,040 & 2,026 & -- \\
        NC4K~\cite{nc4k}            & COS   & 4,121  & --    & 4,121 & -- \\
        \bottomrule
    \end{tabular}
    }
\end{table}

\subsubsection{Evaluation Metrics}
To ensure fair and comprehensive evaluation of OVCOS performance, we adopt a set of evaluation metrics tailored for OVCOS, which are adapted from those originally proposed in the camouflaged scene understanding task~\cite{sinet,r2}. Specifically, we use six metrics: class structure measure \( cS_m \), class weighted F-measure \( cF^{w}_{\beta} \), class mean absolute error \( cMAE \), class standard F-measure \( cF_{\beta} \), class enhanced alignment measure \( cE_m \), and class intersection over union \( cIoU \). These metrics are standard in the open-vocabulary segmentation literature~\cite{r4,r6,r12,r7,r5,r9}, jointly assessing classification accuracy and segmentation quality for a balanced evaluation of model performance.

For the COS task, we follow established protocols~\cite{sinet} and adopt four commonly used metrics: structure measure \( S_\alpha \), enhanced alignment measure \( E_\phi \), weighted F-measure \( F^\omega_\beta \), and mean absolute error \(MAE\). Among the four standard COS metrics, \( S_\alpha \), \( E_\phi \), and \( F^\omega_\beta \) evaluate structural and region-aware similarity between predictions and ground truth, where higher values indicate better performance. Conversely, \(MAE\) measures pixel-wise error, with lower values indicating better accuracy.

\begin{table*}[htbp]
\centering
\caption{Comparison of our method with state-of-the-art CLIP-based open-vocabulary segmentation approaches and the baseline model OVCoser on the OVCamo~\cite{r1} dataset. The bolded values indicate the results of our method, which achieves the best overall performance. The second best is underlined.}
\vspace{1mm}
% \resizebox{\linewidth}{!} % 嵌套环境或自定义宽度环境中
\resizebox{\textwidth}{!} % 主文档环境或浮动体
{
    \begin{tabular}{lccc|cccccc}
        \toprule
        \textbf{Model} &
        VLM & Train Set & Finetune &
        $cS_m \uparrow$ & $cF_{\beta}^w\uparrow$ & $cMAE\downarrow$ & $cF_{\beta}\uparrow$ & $cE_m\uparrow$ & $cIoU\uparrow$ \\
        \midrule
        SimSeg~\cite{r7}   & CLIP-ViT-B/16~\cite{r13} & COCO-Stuff~\cite{r15} & OVCamo~\cite{r1} & 0.098 & 0.071 & 0.852 & 0.081 & 0.128 & 0.0 \\
        OVSeg~\cite{r5}    & CLIP-ViT-L/14~\cite{r13}  & COCO-Stuff~\cite{r15} & OVCamo~\cite{r1} & 0.164 & 0.131 & 0.763 & 0.147 & 0.208 & 0.123 \\
        ODISE~\cite{r11}    & CLIP-ViT-L/14~\cite{r13} & COCO-Stuff~\cite{r15} & OVCamo~\cite{r1} & 0.182 & 0.125 & 0.691 & 0.219 & 0.309 & 0.189 \\
        SAN~\cite{r6}      & CLIP-ViT-L/14~\cite{r13}  & COCO-Stuff~\cite{r15} & OVCamo~\cite{r1} & 0.321 & 0.216 & 0.550 & 0.236 & 0.331 & 0.204 \\
        FC-CLIP~\cite{r12}  & CLIP-ConvNeXt-L~\cite{r14}  & COCO-Stuff~\cite{r15} & OVCamo~\cite{r1}   & 0.124 & 0.074 & 0.798 & 0.088 & 0.162 & 0.072 \\
        CAT-Seg~\cite{r4}  & CLIP-ViT-L/14~\cite{r13}  & COCO-Stuff~\cite{r15} & OVCamo~\cite{r1} & 0.185 & 0.094 & 0.702 & 0.110 & 0.185 & 0.088 \\
        OVCoser~\cite{r1} & CLIP-ConvNeXt-L~\cite{r14}  & OVCamo~\cite{r1} & -- & \underline{0.579} & \underline{0.490} & \underline{0.336} & \underline{0.520} & \underline{0.616} & \underline{0.443} \\
        Ours & Our Fine-Tuned CLIP & OVCamo~\cite{r1} & -- & \textbf{0.668} & \textbf{0.615} & \textbf{0.265} & \textbf{0.631} & \textbf{0.697} & \textbf{0.568} \\
        \bottomrule
    \end{tabular}
}
\label{tab:ovcamo_comparison}
\end{table*}

\begin{table*}[htbp]
\centering
\caption{Quantitative COS task comparison results on three benchmark datasets. The best performance per metric is highlighted in bold, and the second best is underlined. }
\renewcommand{\arraystretch}{1.0}
\resizebox{\textwidth}{!}
{
    \begin{tabular}{l|cccc|cccc|cccc}
    \toprule
    \multirow{2}{*}{Method} & \multicolumn{4}{c|}{CAMO~\cite{camo}} & \multicolumn{4}{c|}{COD10K~\cite{sinet}} & \multicolumn{4}{c}{NC4K~\cite{nc4k}} \\
     & $S_\alpha \uparrow$ & $E_\phi \uparrow$ & $F^{\omega}_\beta \uparrow$ & $MAE \downarrow$
     & $S_\alpha \uparrow$ & $E_\phi \uparrow$ & $F^{\omega}_\beta \uparrow$ & $MAE \downarrow$
     & $S_\alpha \uparrow$ & $E_\phi \uparrow$ & $F^{\omega}_\beta \uparrow$ & $MAE \downarrow$ \\
    \midrule
    SINet~\cite{sinet-0} & 0.751 & 0.771 & 0.606 & 0.100 & 0.771 & 0.806 & 0.551 & 0.051 & 0.808 & 0.883 & 0.768 & 0.058 \\
    RankNet~\cite{nc4k} & 0.712 & 0.791 & 0.583 & 0.104 & 0.767 & 0.861 & 0.611 & 0.045 & 0.840 & 0.904 & 0.802 & 0.048 \\
    PFNet~\cite{pfnet} & 0.782 & 0.852 & 0.695 & 0.085 & 0.800 & 0.868 & 0.660 & 0.040 & 0.829 & 0.887 & 0.784 & 0.053 \\
    SINetV2~\cite{sinet} & 0.820 & 0.882 & 0.743 & 0.070 & 0.815 & 0.887 & 0.680 & 0.037 & 0.847 & 0.903 & 0.770 & 0.048 \\
    ZoomNet~\cite{zoomnet} & 0.820 & 0.892 & 0.752 & 0.066 & 0.838 & 0.911 & 0.729 & 0.029 & 0.853 & 0.912 & 0.784 & 0.043 \\
    SegMaR~\cite{segmar} & 0.815 & 0.872 & 0.742 & 0.071 & 0.833 & 0.895 & 0.724 & 0.033 & 0.841 & 0.905 & 0.781 & 0.046 \\
    DGNet~\cite{dgnet} & 0.839 & 0.901 & 0.769 & 0.057 & 0.822 & 0.896 & 0.693 & 0.033 & 0.857 & 0.911 & 0.784 & 0.042 \\
    SAM~\cite{r20} & 0.684 & 0.687 & 0.606 & 0.132 & 0.783 & 0.798 & 0.701 & 0.050 & 0.767 & 0.776 & 0.696 & 0.078 \\
    SAM-Adapter~\cite{samadapter}& \underline{0.847} & 0.873 & 0.765 & 0.070 & \underline{0.883} & \underline{0.918} & \underline{0.801} & \underline{0.025} & - & - & - & - \\
    MedSAM~\cite{sammed} & 0.820 & \textbf{0.904} & \underline{0.779} & \underline{0.065} & 0.841 & 0.917 & 0.751 & 0.033 & \underline{0.866} & \underline{0.929} & \underline{0.821} & \underline{0.041} \\
    % COMPrompter^{25}\cite{comprompter} & \underline{0.906} & \underline{0.955} & 0.857 & 0.026 & \textbf{0.882} & \textbf{0.942} & \textbf{0.858} & \textbf{0.044} & \underline{0.889} & \textbf{0.949} & \underline{0.821} & \underline{0.023} & \textbf{0.907} & \textbf{0.955} & \textbf{0.876} & \textbf{0.030} \\
    Ours & \textbf{0.865} & \underline{0.902} & \textbf{0.789} & \textbf{0.057} & \textbf{0.905} & \textbf{0.947} & \textbf{0.845} & \textbf{0.019} & \textbf{0.904} & \textbf{0.933} & \textbf{0.852} & \textbf{0.031} \\
    \bottomrule
    \end{tabular}%
}
\label{tab:cos_results}
\end{table*}

\subsubsection{Training Details}

All experiments are conducted on a system equipped with two NVIDIA RTX 3090Ti GPUs running Ubuntu 20.04. Our framework is implemented on PyTorch with CUDA 11.8 acceleration.

During the CLIP model fine-tuning, we adopt a multi-modal prompting strategy on the pre-trained ViT-L/14 Alpha-CLIP model~\cite{r21}. The model is trained on the OVCamo~\cite{r1} dataset for 10 epochs using stochastic gradient descent (SGD) with a learning rate of 0.0035 and a batch size of 8 on a single GPU, following the setup in~\cite{r10}. Additionally, the input alpha mask is randomly selected as either an all-one mask or the ground truth segmentation mask with equal probability, balancing global context encoding and localized focus.

During adapted SAM training, the fine-tuned CLIP is integrated into our adapted SAM architecture, based on the ViT-H variant of SAM~\cite{r20}. The network is trained for 20 epochs using the Adam optimizer with an initial learning rate of \( 2 \times 10^{-4} \), decayed via cosine annealing. Training is conducted on two GPUs with a batch size of 2 and completes in approximately 24 hours.

\subsection{Experimental Results}

\begin{figure*}[!htb]
  \centering
  \includegraphics[width=\linewidth]{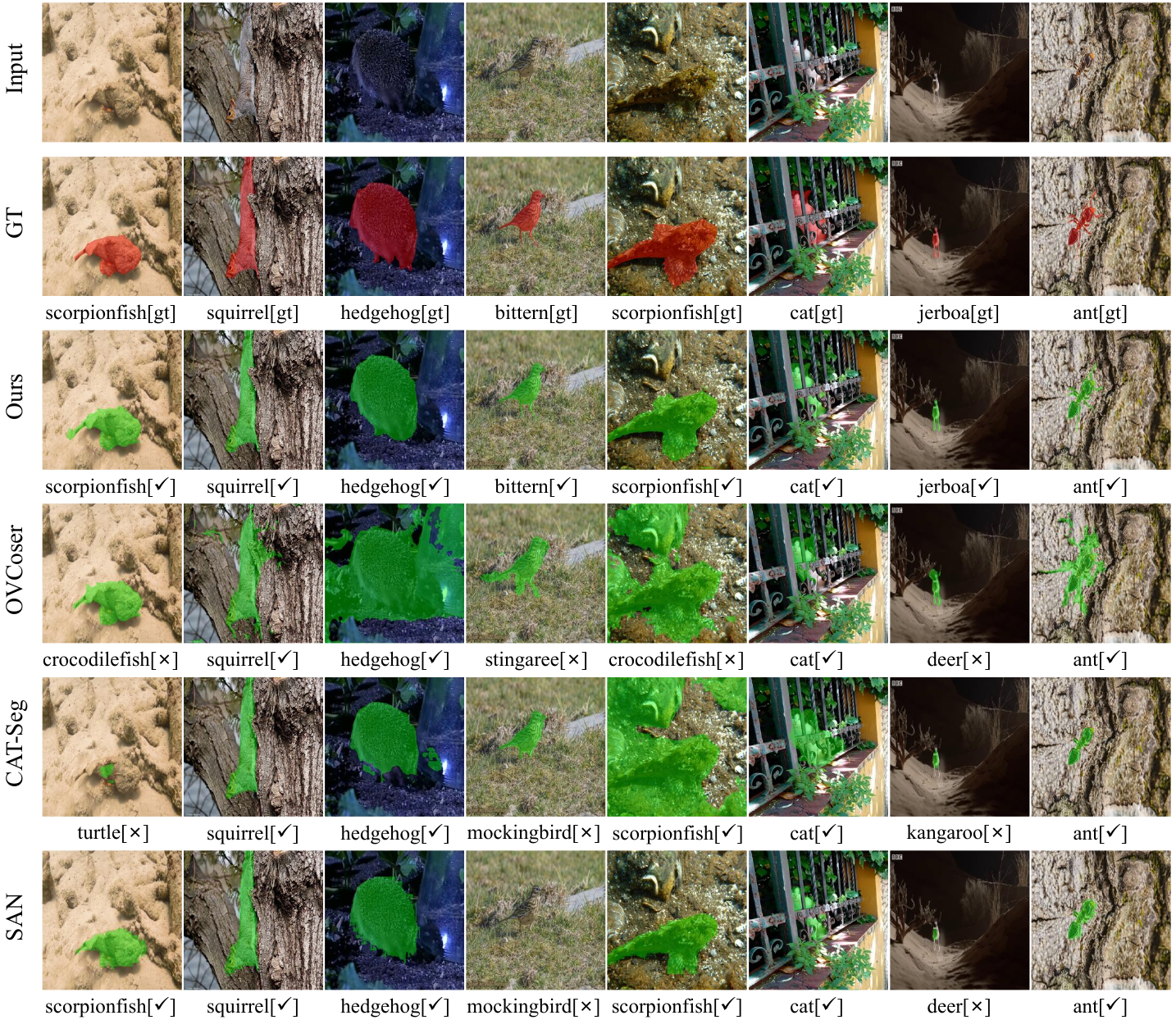}
  \caption{Qualitative comparison between our method and CLIP-based baselines on OVCamo~\cite{r1}. Each column depicts the input image, segmentation result, and predicted label. Predicted label is shown below each segmentation result, where [\cmark] indicates correct prediction and [\xmark] denotes an incorrect one.}
  \label{fig:viscmp}
  % \vspace{-1em}
\end{figure*}

\label{sec:experimental_results}
\subsubsection{Quantitative Results on OVCOS}

To thoroughly evaluate the effectiveness of our proposed framework, we compare it with recent state-of-the-art open-vocabulary segmentation methods, including CAT-Seg~\cite{r4}, SAN~\cite{r6}, SimSeg~\cite{r7}, OVSeg~\cite{r5}, FC-CLIP~\cite{r12}, ODISE~\cite{r11} and the baseline OVCoser~\cite{r1}. For a fair comparison, all models are trained or fine-tuned on the OVCamo~\cite{r1} dataset. We adopt the large variants of such approaches wherever available, except for SimSeg, which is only released in its base form. As shown in Table~\ref{tab:ovcamo_comparison}, our method consistently outperforms all competitors across multiple evaluation metrics.

While open-vocabulary segmentation methods such as SAN~\cite{r6}, OVSeg~\cite{r5}, and CAT-Seg~\cite{r4} benefit from large-scale pretraining, they lack task-specific adaptation, resulting in limited performance on OVCOS (e.g., OVSeg: 0.164 $cS_m$, 0.123 $cIoU$). The baseline OVCoser~\cite{r1} improves results by integrating camouflage segmentation with CLIP-based classification, achieving 0.579 $cS_m$ and 0.443 $cIoU$, but it does not fine-tune vision-language embeddings or incorporate semantic guidance into segmentation.

Compared to the results of existing methods shown in Table~\ref{tab:ovcamo_comparison}, our framework leverages fine-tuned CLIP and a task-adapted SAM to enhance both segmentation and classification. Ours achieves state-of-the-art results, surpassing the baseline OVCoser~\cite{r1} by notable margins across all metrics: +8.9\% in $cS_m$, +12.5\% in $cIoU$, +12.5\% in $cF_{\beta}^w$, +11.1\% in $cF_{\beta}$, +8.1\% in $cE_m$, and a reduction of 7.1\% in $cMAE$. These results highlight the effectiveness of our cascaded design and cross-modal semantic conditioning in tackling the OVCOS challenge.

\begin{table}[!htb]
    \centering
    \caption{Classification performance comparison of different CLIP models on the OVCamo~\cite{r1} test set.}
    \small % 或 \scriptsize
    \begin{tabularx}
    {\linewidth}{l>{\centering\arraybackslash}X>{\centering\arraybackslash}X>{\centering\arraybackslash}X}
        \toprule
        \textbf{Model} & \textbf{Alpha} & \textbf{Top-1}$\uparrow$ & \textbf{Top-5}$\uparrow$ \\
        \midrule
        CLIP-ConvNeXt-L~\cite{r14} & -- & 0.6944 & 0.8918 \\
        CLIP-ViT-L/14~\cite{r13} & -- & 0.7040 & 0.8915 \\
        Alpha-CLIP~\cite{r21} & all one & 0.6934 & 0.8849 \\
        Alpha-CLIP~\cite{r21} & gt & 0.7467 & 0.9456 \\
        Ours & all one & 0.7462 & 0.9003 \\
        Ours & gt & \textbf{0.7859} & \textbf{0.9497} \\
        \bottomrule
    \end{tabularx}
    \label{tab:classification_alpha_clip}
\end{table}

\subsubsection{Quantitative Results on COS}

As shown in Table~\ref{tab:cos_results}, our adapted SAM model achieves competitive performance across three widely used COS benchmarks: CAMO~\cite{camo}, COD10K~\cite{sinet}, and NC4K~\cite{nc4k}. Compared to both traditional non-SAM-based methods~\cite{sinet, sinet-0, nc4k, pfnet, zoomnet, dgnet, segmar} and recent SAM-based approaches~\cite{r20, samadapter, sammed}, our model consistently outperforms across all datasets.

Specifically, the adapted SAM ranks first on 11 out of 12 evaluation metrics and second on the remaining one, demonstrating strong generalization and robustness in diverse camouflage scenarios. Ours achieves notable improvements in structure-aware metrics ($S_\alpha$, $E_\phi$), region-aware precision ($F^\omega_\beta$), and pixel-level accuracy (MAE), particularly on the COD10K and NC4K datasets. These results highlight the effectiveness of our edge-enhanced architecture and prompt-guided segmentation in capturing fine-grained boundaries and ensuring semantic consistency.

\subsubsection{Qualitative Results of OVCOS}
To further validate our quantitative findings, we present qualitative comparisons in Figure~\ref{fig:viscmp}. Ours consistently delivers superior segmentation quality, accurately delineating camouflaged objects with well-preserved shapes and precise boundaries—even in low-contrast and cluttered backgrounds. Compared to other methods, our approach better maintains object integrity and minimizes background leakage, demonstrating enhanced robustness in challenging camouflage scenarios.

In terms of classification, ours reliably predicts correct categories across diverse samples, outperforming prior methods that often misclassify visually ambiguous targets. This classification accuracy improvement stems from our region-aware classification strategy, which integrates segmentation masks as alpha masks into the fine-tuned CLIP model. Combined with multi-modal prompting and edge-aware decoding, ours achieves high fidelity in both localization and recognition under open-vocabulary conditions.

\begin{table*}[htbp]
    \centering
    \caption{Ablation results showing the effectiveness of our fine-tuned CLIP on OVCOS performance.}
    \label{tab:alpha_clip_ablation}
    \begin{minipage}{0.8\textwidth} % 控制表格宽度，例如 0.85
    % \small
    \begin{tabularx}{\linewidth}{Xccccccc}
        \toprule
        \textbf{Model} & \textbf{VLM} & \({cS_m\uparrow}\) & \({cF_{\beta}^{w}\uparrow}\) & \({cMAE}\downarrow\) & \({cF_{\beta}\uparrow}\) & \({cE_m}\uparrow\) & \({cIoU}\uparrow\) \\
        \midrule
        COCUS & CLIP-ConvNeXt-L~\cite{r14} & 0.567 & 0.518 & 0.375 & 0.534 & 0.591 & 0.481 \\
        COCUS & CLIP-ViT-L/14~\cite{r13} & 0.580 & 0.536 & 0.353 & 0.551 & 0.617 & 0.503 \\
        COCUS & Alpha-CLIP~\cite{r21} & 0.639 & 0.589 & 0.299 & 0.603 & 0.668 & 0.545 \\
        COCUS & Our Fine-Tuned CLIP & \textbf{0.668} & \textbf{0.615} & \textbf{0.265} & \textbf{0.631} & \textbf{0.697} & \textbf{0.568} \\
        \bottomrule
    \end{tabularx}
    \end{minipage}
\end{table*}

\begin{table}[htbp]
    \centering
    \footnotesize  % 字体稍微小一些，避免数字太宽
    \caption{Ablation study of Conditional Multi-Way Attention (CMA) and Edge Enhancement (EDE) in the adapted mask decoder.}
    \label{tab:ablation_cma_edc}
    \begin{tabularx}{0.52\linewidth}{
        >{\raggedright\arraybackslash}p{1.6cm}  % 模型列
        >{\centering\arraybackslash}p{0.6cm}    % 指标列
        >{\centering\arraybackslash}p{0.6cm}
        >{\centering\arraybackslash}p{0.9cm}
        >{\centering\arraybackslash}p{0.6cm}
        >{\centering\arraybackslash}p{0.6cm}
        >{\centering\arraybackslash}p{0.6cm}
    }
        \toprule
        \textbf{Model} & $cS_m\uparrow$ & $cF_{\beta}^{w}\uparrow$ & $cMAE\downarrow$ & $cF_{\beta}\uparrow$ & $cE_m\uparrow$ & $cIoU\uparrow$ \\
        \midrule
        Baseline         & 0.644 & 0.599 & 0.281 & 0.610 & 0.651 & 0.549 \\
        + EDE            & 0.650 & 0.605 & 0.278 & 0.615 & 0.666 & 0.554 \\
        + CMA            & 0.652 & 0.607 & 0.273 & 0.621 & 0.683 & 0.551 \\
        + CMA, EDE       & \textbf{0.668} & \textbf{0.615} & \textbf{0.265} & \textbf{0.631} & \textbf{0.697} & \textbf{0.568} \\
        \bottomrule
    \end{tabularx}
\end{table}
\subsection{Ablation Study}
\label{sec:ablation_results}
\subsubsection{Effectiveness of the Fine-Tuned CLIP}

To assess the effectiveness of our fine-tuned CLIP model, we conduct comparative experiments on classification accuracy using several CLIP~\cite{r13} variants, including CLIP-ConvNeXt-L~\cite{r14}, CLIP-ViT-L/14~\cite{r13}, the original Alpha-CLIP~\cite{r21}, and ours. Evaluation results on the OVCamo~\cite{r1} test set are presented in Table~\ref{tab:classification_alpha_clip}, where \textit{all one} refers to using an all-one alpha mask, and \textit{gt} denotes the ground-truth segmentation mask.

The results clearly show that Alpha-CLIP~\cite{r21} with \textit{gt} alpha mask significantly outperforms both CLIP-ConvNeXt-L~\cite{r14} and CLIP-ViT-L/14~\cite{r13}, the backbone models used in OVCoser~\cite{r1}, demonstrating the advantage of incorporating an auxiliary alpha mask for open-vocabulary classification. Moreover, ours further improves classification accuracy, reducing the performance gap between the \textit{all one} and \textit{gt} settings. Ours achieves the highest accuracy when using the \textit{gt} mask, emphasizing the benefit of integrating task-specific semantic information during inference.

Additionally, we evaluate the impact of integrating the fine-tuned CLIP into ours framework. As shown in Table~\ref{tab:alpha_clip_ablation}, our fine-tuned CLIP model consistently outperforms the original Alpha-CLIP across all key metrics, confirming its effectiveness in enhancing semantic representation and improving overall model performance on the OVCOS task.

\subsubsection{Impact of Adapted Mask Decoder}\label{sec:abl-mask-dec}

As shown in Table~\ref{tab:ablation_cma_edc}, we perform ablation studies on the OVCamo~\cite{r1} dataset to assess the effectiveness of the proposed Conditional Multi-Way Attention (CMA) and Edge Enhancement (EDE) modules in our adapted mask decoder. Beginning with a baseline SAM~\cite{r20} model equipped with lightweight adapters—corresponding to the original SAM mask decoder without either enhancement—we observe consistent performance improvements when incorporating CMA or EDE individually.

Specifically, adding the EDE module (\textit{+ EDE}) leads to notable gains in contour-sensitive metrics such as \(cF_{\beta}^{w}\) and \(cIoU\), indicating that explicit edge modeling enhances boundary precision. In contrast, introducing the CMA module (\textit{+ CMA}) results in stronger improvements in semantic-aware metrics like \(cE_m\) and \(cF_{\beta}\), demonstrating that conditional attention effectively enriches textual-visual feature fusion.

When both modules are combined, our method achieves the best performance across all metrics, underscoring the complementary strengths of semantic conditioning and edge-aware refinement. These findings confirm the importance of both enhancements in improving segmentation quality for challenging camouflaged object scenarios.

\section{Conclusion}

In this paper, 
we present COCUS, 
a two-stage framework for OVCOS that explicitly 
decouples segmentation and classification.
In the first stage, visual and textual embeddings are extracted using our fine-tuned CLIP model. These embeddings guide an adapted SAM with a redesigned mask decoder to enhance object localization and boundary precision. In the second stage, the predicted segmentation mask is fused with the input image to guide the attention of the model toward the target regions, enabling region-aware classification without relying on cropped inputs. Extensive experiments on both OVCOS and COS benchmarks show that ours outperforms existing open-vocabulary segmentation methods. 
The adapted SAM also achieves superior results on the COS benchmarks. These experiments confirm the benefits of our two-stage framework and edge-aware
 enhancements in complex camouflage scenarios.

\bibliographystyle{plain}
\bibliography{reference}

\end{document}